\documentclass{article}
\usepackage{spconf,amsmath,graphicx}
\usepackage{times}
\usepackage{epsfig}
\usepackage{amssymb}
\usepackage{amsfonts} 
\usepackage[normalem]{ulem}
\usepackage[table,xcdraw,dvipsnames]{xcolor}
\usepackage{colortbl}
\usepackage{stfloats}

\def\etc{{etc\ }}

\usepackage{color,soul}  
\definecolor{ao(english)}{rgb}{0.0, 0.5, 0.0}

\title{Multi-stage Based Feature Fusion of Multi-Modal Data for Human Activity Recognition}
\name{Hyeongju Choi \qquad Apoorva Beedu \qquad Harish Haresamudram \qquad Irfan Essa
\vspace{-0.1in}}

\address{
\vspace{-0.02in}
Georgia Institute of Technology, USA \\ \{hchoi375, abeedu3, hharesamudram3, irfan\}@gatech.edu
\vspace{-0.15in}}

\begin{document}
%
\maketitle

\begin{abstract}
To properly assist humans in their needs, human activity recognition (HAR) systems need the ability to fuse information from multiple modalities. 
Our hypothesis is that multimodal sensors, visual and non-visual tend to provide complementary information, addressing the limitations of other modalities.  
In this work, we propose a multi-modal framework that learns to effectively combine features from RGB Video and IMU sensors, and show its robustness for MMAct and UTD-MHAD datasets.
Our model is trained in two-stage, where in the first stage, each input encoder learns to effectively extract features, and in the second stage, learns to combine these individual features. We show significant improvements of 22\% and 11\% compared to video only and IMU only setup on UTD-MHAD dataset, and ~20\% and 12\% on MMAct datasets.
Through extensive experimentation, we show the robustness of our model on zero shot setting, and limited annotated data setting. 
We further compare with state-of-the-art methods that use more input modalities and show that our method outperforms significantly on the more difficult MMact dataset, and performs comparably in UTD-MHAD dataset.
\end{abstract}
%
\setlength{\skip\footins}{0.3cm}
\let\thefootnote\relax\footnote{Under review}

\vspace{-0.1in}
\section{Introduction}
\label{sec:intro}

In recent years, as we have seen a rapid increase in the bodily worn sensors and it's applications from tracking GPS to automatically switching between activities e.g. triathlons, we have seen a boom in the machine learning algorithms that combine information from multiple sensors for Human Activity Recognition (HAR). 
With this rise of bodily worn sensors and smart home systems~\cite{mehr2019human,du2019novel}, we have seen the expansion to vision-based HAR applications in human-robot interaction, senior care, healthcare, personal fitness, and surveillance~\cite{zeng, zhang2017review}. Thus depending on the type of system used, different input modalities such as RGB-D, IMU sensors, IoT device information \etc can be used for HAR. 
Using only IMU data limits the complexity and variety of the actions and activities that can be detected and classified by the HAR system. 
For instance, the HAR system that relies only on IMU data from the inertial sensor equipped on a wrist might not be able to distinguish between waving and cleaning a window. 
Hence, these limitations call for a multimodal approach that can leverage other data for accurate detections.

Multimodal HAR is a paradigm that uses data from more than one modality (e.g., RGB videos, IMU sensors, Skeleton joints, or depth videos) and learns an effective way to combine these different modalities to improve the accuracy of the HAR system.
With this motivation, our contributions are as follows:
\emph{(1)} We propose a HAR framework that works on multi-modal inputs such as RGB video and IMU data.
\emph{(2)} We compare single modality vs multi-modal input performance for two datasets: UTD-MHAD \cite{UTD-MHAD} and MMAct \cite{mmact}.
\emph{(3)} Through extensive experimentation, we show the effectiveness of our framework even when limited data is available and in zero-shot setting.

\begin{figure*}[t]
\centering
    \includegraphics[width=0.85\textwidth]{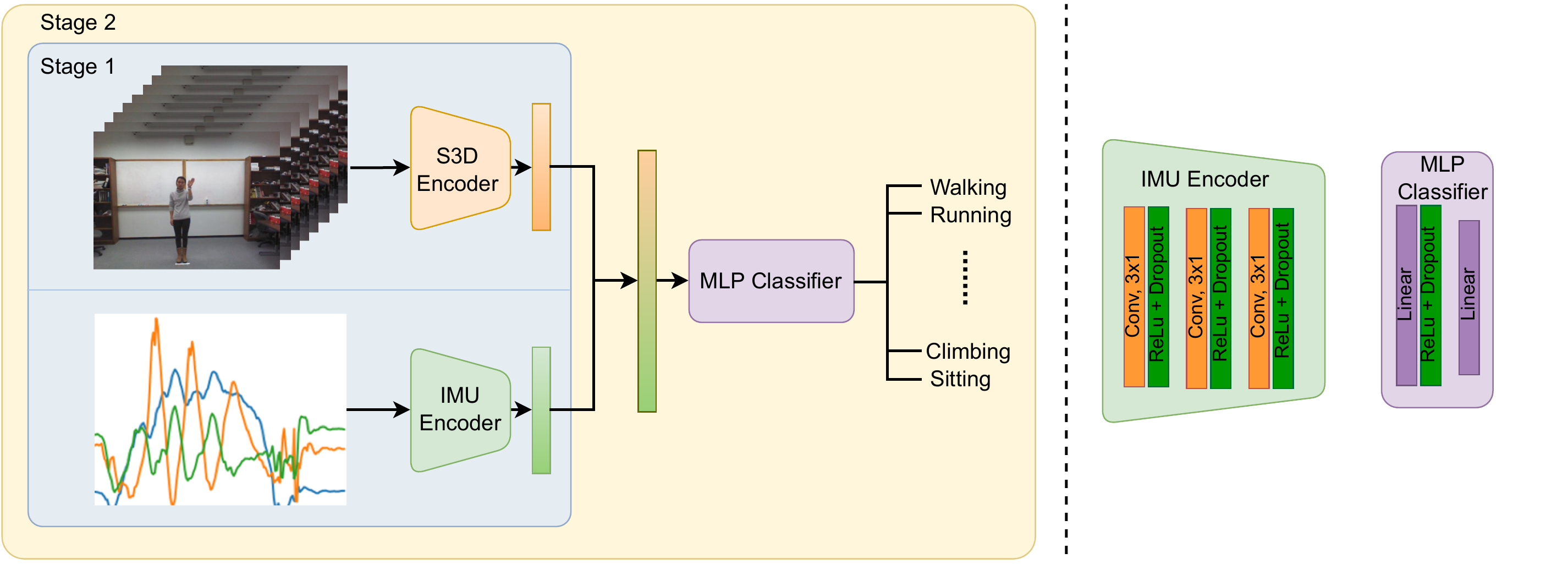}
\label{fig:model_framework}
\caption{\textbf{\textit{Left:}} Overview of our framework. Videos are encoded using an 3D-CNN architecture: S3D, and IMU data are encoded using  3 blocks of 1D-CNN layers. These features are then fused and passed through a two layer MLP to predict action classes. \textbf{\textit{Right:}} IMU Encoder with three layers of 1D convolution + Dropout + ReLu, and MLP Classifier with two linear layer, and one non-linear activation between them.
\vspace{-0.1in}}
\end{figure*}

\vspace{-0.1in}
\section{Related Work}
\vspace{-0.05in}
\label{sec:related}
Our work is focused on learning strong representations for the task of human activity recognition. In what follows, we divide the existing literature into two categories based on the number of input modalities used and summarize these works.
\vspace{-0.3in}
\subsection{Uni-modal Human Activity Recognition}
\vspace{-0.1in}
Single modality based HAR systems have been extensively studied by both ubiquitous computing and vision communities ~\cite{j8, j9, rahmani20163d,liu2017skeleton,donahue2015long,mehr2019human}.
Vision-based activity recognition's effectiveness mainly stems from the expanding field of representation learning for image and video features\cite{liu2017skeleton,j9,I3d}. 
In addition, the abundance of annotated data from datasets like Kinetics~\cite{kinetics}, HowTo100M~\cite{j9} has also benefited the vision-based HAR systems. 
However, the vision-based systems are typically affected by the recording conditions (camera angles, viewpoint, lighting, background variations etc.) as well as occlusions.
For target scenarios where privacy is a concern, e.g., medical applications, obtaining recorded video may be very challenging, or even impossible.
In such scenarios, inertial measurement unit (IMU) signals are more suitable, as they are generally available on commodity platforms including smartphones and smartwatches and can record data without disruption of user experience.
Due to this, recent years have seen significant interest in developing sensor-based activity recognition systems \cite{hammerla2016deep, murahari2018attention, haresamudram2021contrastive}.  
Although the single modality based approaches have demonstrated their effectiveness in recognizing activities, each modality has weaknesses that the other can complement. 
For example, sensitivity to occlusions can be overcome by using IMU data, and the sensitivity of IMU sensors to body-worn positions~\cite{mukhopadhyay2014wearable} can be addressed by using videos and skeleton data. 
\vspace{-0.15in}
\subsection{Multi-modal Human Activity Recognition}
\vspace{-0.1in}
The advantages of complementary modalities have been explored in computer vision for a wide range of tasks. 
For example, object pose estimation methods often use RGB + Depth data \cite{wang2019densefusion,wang2022phocal}, whereas Visual Question Answering (VQA) utilizes text questions on RGB images for answering \cite{visdial,vqa}. 
Action Recognition tasks have also combined inputs from multiple modalities~\cite{ardianto2018multi} like depth and infrared data in addition to RGB videos to estimate predefined human actions.
With the introduction of Transformer \cite{vaswani2017attention} based methods, works such as UniVL~\cite{luo2020univl}, Omnivore~\cite{girdhar2022omnivore}, HAMLET~\cite{islam2020hamlet} effectively learn to fuse information from multiple modalities for the task of action recognition. UniVL~\cite{luo2020univl} combines information from RGB videos and textual inputs to answer questions about a video and generate captions. Omnivore~\cite{girdhar2022omnivore} on the other hand learns to effectively learn joint representation from multiple input modalities such as RGB images, videos, and depth maps. This learnt  shared representation is then fine-tuned for action and image classification tasks.
However, the combination of Video and IMU data has seen limited exploration, though recently, techniques such as~\cite{islam2022maven,ni2022cross,islam2022mumu} have been introduced.
However, these methods generally use skeleton data, or perform fusion and learn features in a non-synchronous way. 
In contrast, our framework specifically works on the synchronized RGB Video and IMU data, and is trained in a two-stage method that allows for the model to learn strong representations from individual input modalities in the first stage, and effectively fuse the information in the second stage.
\vspace{-0.1in}
\section{Methodology}
\label{sec:method}
\vspace{-0.05in}
To address the problems of single modality HAR systems, we introduce a multimodal framework using two modalities: third-person view RGB video and IMU data. 
We discuss the different feature-extracting modules and fusion of these different features obtained in this section. 
An overview of the network is shown in Figure~\ref{fig:model_framework}.


\vspace{-0.13in}
\subsection{Feature Encoder for IMU data}
\vspace{-0.1in}
\label{sec:feature_encoder_imu}
IMU data is a multivariate time-series data that (generally) contains data from accelerometers, gyroscopes, and sometimes magnetometers on the x, y, and z axes. 
1D Convolutional neural networks (CNNs) are known to be effective for feature learning, as they can parse data across time, thereby capturing the time-series nature of sensor data \cite{j8}. 
We implement a simple encoder for the IMU data, similar to~\cite{zeng2014convolutional}, but comprising 3 blocks of 1D convolutional layers. 
The kernel size is set to 24, 16, and 8 respectively, and the architecture is shown in Figure ~\ref{fig:model_framework}. 
The IMU encoder is initialized with random weights during training. 
\vspace{-0.13in}
\subsection{Feature Encoder for Video}
\vspace{-0.1in}
\label{sec:feature_encoder_video}
We use the S$3$D network~\cite{j9} to encode the video features. 
The S$3$D network is a video classification network that projects a video into a temporal and spatial representation and is pre-trained on Kinetics400\cite{kinetics}.
\vspace{-0.1in}
\subsection{End-to-End training}
\vspace{-0.1in}
Our framework is trained in two stages. In the first stage, we pre-train both the video and the IMU encoders separately as a supervised action recognition task. 
As a second stage, we fine-tune the \textit{Mixed\_4c} and \textit{Mixed\_5c} layers of the S$3$D architecture, and all the layers of the IMU encoder to predict action classes.
Features from both the encoders are concatenated together and then passed through a 2 layer MLP structure to predict the final action classes.

The entire model is trained with a cross-entropy loss for action classification. Details of the implementation are discussed in Section~\ref{sec:implementation}.

\vspace{-0.14in}
\section{Experiments and Results}
\label{sec:results}
\vspace{-0.05in}
\subsection{Dataset} \label{sec:Dataset}
\vspace{-0.1in}
To evaluate the performance of our proposed model in this project, UTD-MHAD \cite{UTD-MHAD} and MMAct \cite{mmact} are used. 

\noindent \textbf{UTD-MHAD dataset} contains a total of 861 samples collected using one wearable inertial sensor and one Kinect camera from 8 subjects, 4 males and 4 females, performing 27 different activities, repeated 4 times for each activity. The dataset consists of 4 different modalities including RGB videos, Depth videos, Skeleton positions, and inertial signals. For our study, only the RGB videos and inertial signals were used. Same as the original paper, we use odd-numbered subjects: 1, 3, 5, 7 for training and even-numbered subjects: 2, 4, 6, 8 for testing for all models. 

\noindent \textbf{MMAct dataset} contains a total of 36764 samples from 37 activities with each activity repeated 5 times by 20 subjects in 4 different scenes including the scene of occlusion to overcome the weakness of the vision-based HAR systems. The dataset consists of 7 modalities including RGB videos and inertial signals. we follow the cross-subject evaluation protocol by using the samples from the first 16 subjects for training and the rest for testing.

\vspace{-0.1in}
\subsection{Implementation}
\vspace{-0.1in}
\label{sec:implementation}
Our model was implemented using the PyTorch~\cite{paszke2019pytorch} framework. Training hyperparameters are detailed in Table~\ref{table:implementation}. All the training for UTD-MHAD was done on a single A40 GPU, while MMAct was trained on 4 A40 GPUs.

\begin{table}
\setlength{\tabcolsep}{6pt} 
\renewcommand{\arraystretch}{1.1}
\centering
\caption {Training hyperparameter details}
\label{table:implementation}
\resizebox{1\columnwidth}{!}{
\begin{tabular}{|c|cccccccc|}
\hline
 &  & \multicolumn{3}{c}{UTD-MHAD} &  & \multicolumn{3}{c|}{MMAct} \\ \cline{3-5} \cline{7-9} 
 &  & lr & \begin{tabular}[c]{@{}c@{}}Weight decay\end{tabular} & batch &  & lr & \begin{tabular}[c]{@{}c@{}}Weight decay\end{tabular} & batch \\ \cline{1-1} \cline{3-5} \cline{7-9} 
IMU &  & 1e-4 & 1e-6 & 128 &  & 5e-3 & 5e-5 & 256 \\
Video &  & 1e-3 & 1e-5 & 16 &  & 1e-3 & 1e-5 & 20 \\
IMU + Video &  & 5e-4 & 5e-6 & 16 &  & 1e-4 & 1e-6 & 18 \\
\hline
\end{tabular}}
\vspace{-0.2in}
\end{table}

\noindent \textbf{Pre-Processing:}
Both RGB videos and IMU signals were pre-processed for feature encoders. RGB frames were sampled at 15HZ for UTD-MHAD and 30 fps for MMAct. The IMU signals were sampled at 50HZ for UTD-MHAD and a mixture of 100HZ (acceleration) and 50HZ (gyroscope, orientation) for MMAct. The signals were padded with zeros whenever a dimension mismatch occurred.

\vspace{-0.15in}
\subsection{Evaluation Metrics}
\vspace{-0.1in}
\label{sec:evaluation_metrics}
To compare the performance of these different HAR models we report Top 1 accuracy, Top 5 accuracy, and F1 score on the test set for each of the datasets. 


\begin{table*}[t]
\setlength{\tabcolsep}{6pt} 
\renewcommand{\arraystretch}{1.1}
\centering
\caption{Action recognition performance for Top-1, Top-5 and F1 score compared with baseline methods.} 
\label{table:baselines}
\resizebox{0.8\textwidth}{!}{
\begin{tabular}{|c|c|ccccccccc|}
\hline
 &  &  & \multicolumn{3}{c}{UTD-MHAD} &  & \multicolumn{3}{c}{MMAct} &  \\ \cline{1-2} \cline{4-6} \cline{8-10}
Method & Data used &  & Top 1 & Top 5 & F1 &  & Top 1 & Top 5 & F1 &  \\ \cline{1-2} \cline{4-6} \cline{8-10}
HAMLET~\cite{islam2020hamlet} & Video  + Skeleton + Sensor &  & 95.12 & - & - &  & - & - & - &  \\
MuMu~\cite{islam2022mumu} & Video + Depth  + Skeleton + Sensor &  & 97.6 & - & - &  & - & - & 76.28 &  \\
CMC-CMKM~\cite{brinzea2022contrastive} & Skeleton + Sensor &  & \textbf{97.21} & - & - &  & \textbf{84.05} & - & - &  \\
VSKD + DASK~\cite{ni2022cross} & Video + Sensor &  & 96.97 & - & \textbf{96.38} &  & - & - & 65.83 &  \\
VSKD ~\cite{ni2022cross} & Sensor &  & 94.87 & - & - &  & - & - & 61.38 &  \\\cline{1-2} \cline{4-10}
Ours (IMU only) & Sensor &  & 85.42 & 97.14 & 85.44 &  & 64.28 & 90.57 & 65.45 &  \\
Ours (Video only) & Video &  & 74.76 & 95.14 & 74.09 &  & 71.89 & 94.92 & 72.86 &  \\
Ours (IMU + Video) & Video + Sensor &  & 96.05 & \textbf{100} & 96.05 &  & \textbf{84.05} & \textbf{98.96} & \textbf{84.78} &  \\ \hline
\end{tabular}
}
\end{table*}

\vspace{-0.1in}
\subsection{Results and Discussions}
\vspace{-0.1in}
We compare the effectiveness of our model with single modality inputs in Table~\ref{table:baselines}. We also compare our results against HAMLET~\cite{islam2020hamlet}, MuMu~\cite{islam2022mumu}, a contrastive learning based method CMC-CMKM~\cite{brinzea2022contrastive}, and VSKD~\cite{ni2022cross}.

We see that when compared against single modality frameworks, our methods perform significantly better (10.6\% compared to IMU, and 21.29\% compared to Video) across all metrics.
When compared to other baseline methods, our method sees a slight drop in performance for UTD-MHAD dataset, but performs comparable or better for MMAct. 
It is worth noting that other methods use Skeleton information in addition to RGB Video and Sensor data. CMC-CMKM, a contrastive loss based framework uses Skeleton in lieu of video, however, this comes at an additional computation cost and speed, as having an accurate skeleton estimation becomes crucial to their method. As opposed to that, our methods use video features directly, without needing a post-processing framework.
However, utilizing skeleton aids in the action recognition task, and we intend to explore this modality further in future work.
Although ~\cite{ni2022cross} uses the same modalities as our framework, they propose a novel loss called DASK that is responsible for the performance improvement. 
Their student network that uses IMU data uses a ResNet-18 based model and performs significantly better than our 1D-CNN based model. 

 


\vspace{-0.1in}
\subsection{Effect of the size of training dataset}
\begin{figure}
\centering
\centering
    \includegraphics[width=0.9\columnwidth]{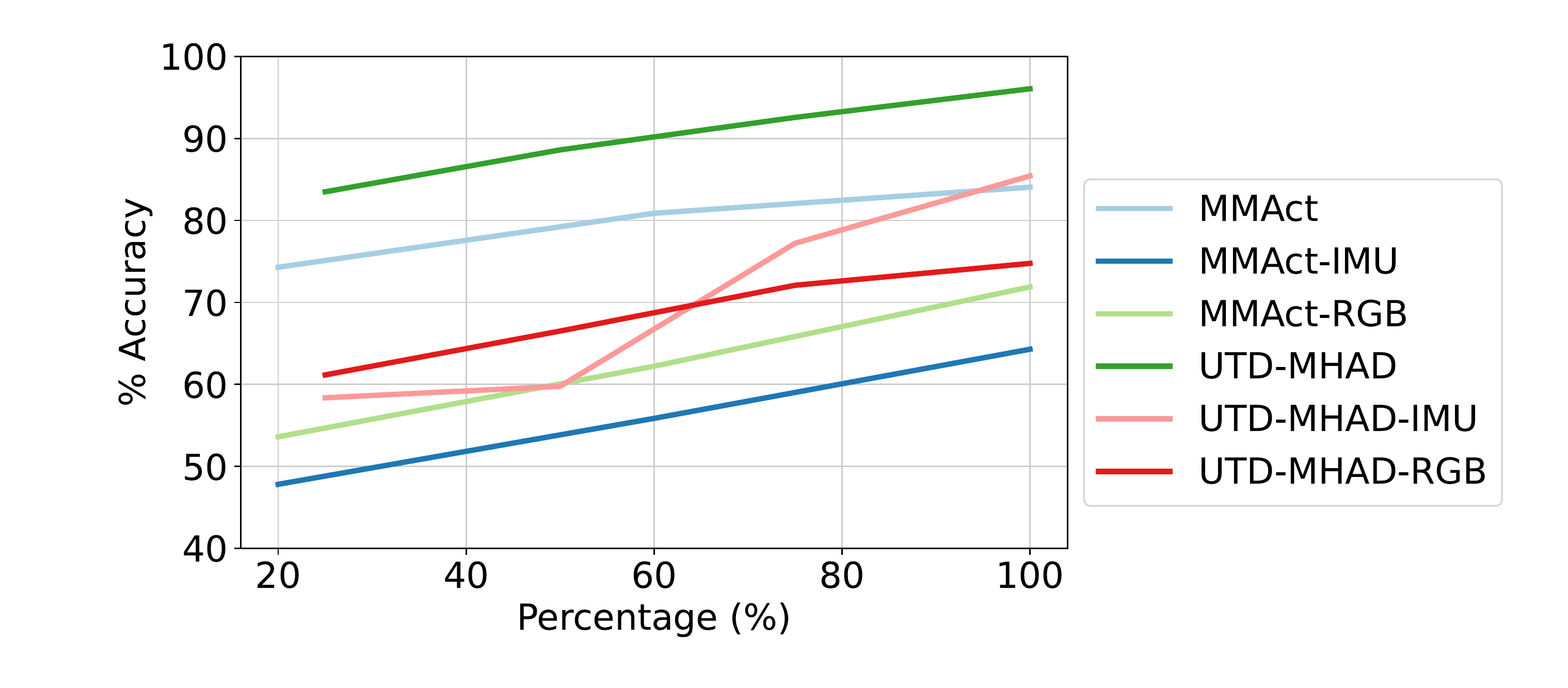}
 \vspace{-0.2in}
\caption{Action recognition performance for different ratio of training data for MMAct and UTD-MHAD datasets.}
 \label{fig:percentage_data_performance}
 \vspace{-0.2in}
\end{figure}

\vspace{-0.1in}
In this experiment, we evaluate the effect of different ratios of the training data on performance. As can be seen in Table~\ref{fig:percentage_data_performance}, we only see a drop of about 10\% on UTD-MHAD and 10\% on MMAct when the training sample is reduced to 25\%. Compared to single modality performances, multi-modal method performs significantly better. This further shows the effectiveness of multi-modal learning for human activity recognition.

\vspace{-0.3in}
\subsection{Zero Shot setting}
\begin{figure*}[t]
\centering
\begin{minipage}[c]{0.49\textwidth}
\centering
    \includegraphics[width=0.8\columnwidth]{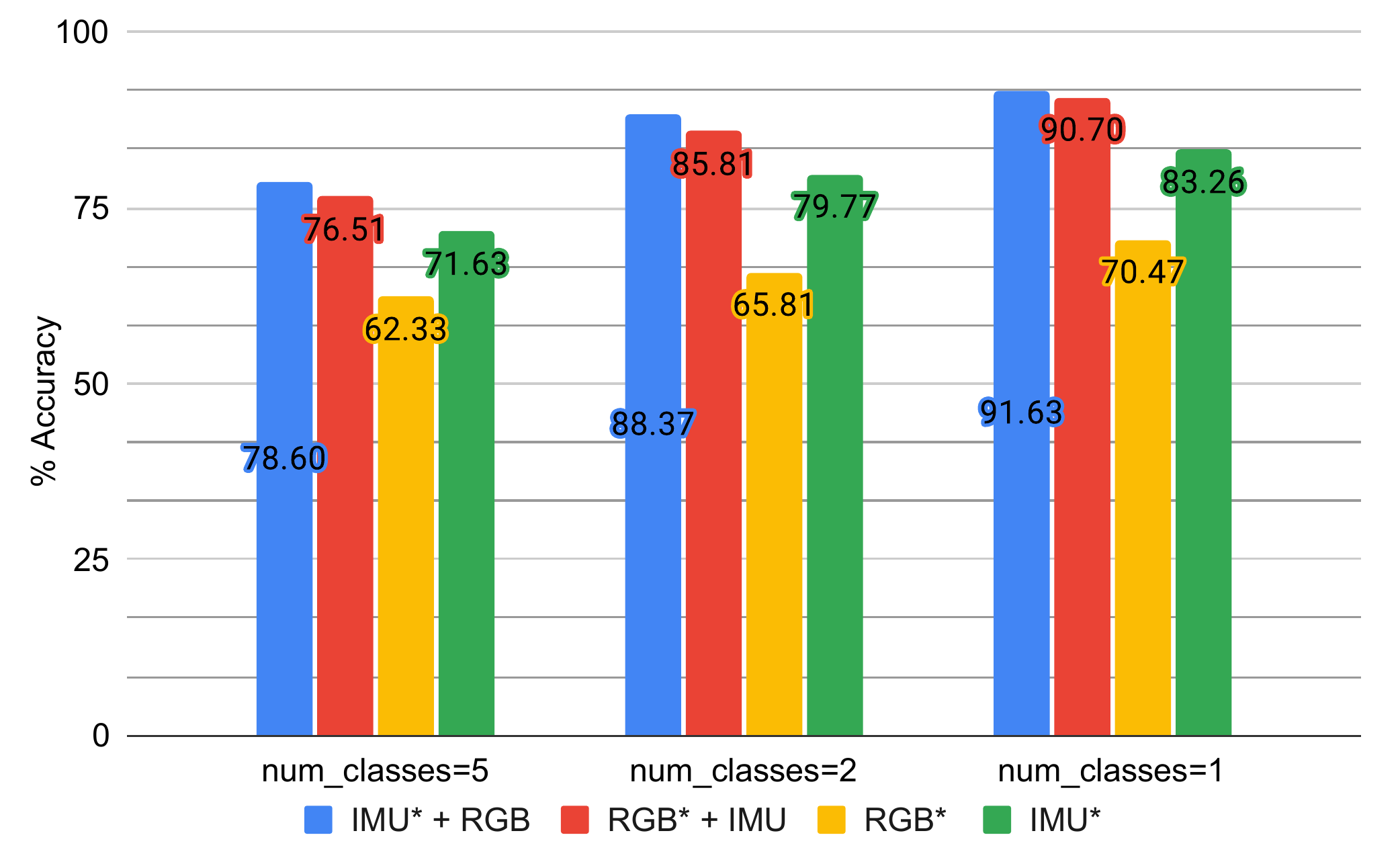}
\end{minipage}
\begin{minipage}[c]{0.49\textwidth}
\centering
    \includegraphics[width=0.8\columnwidth]{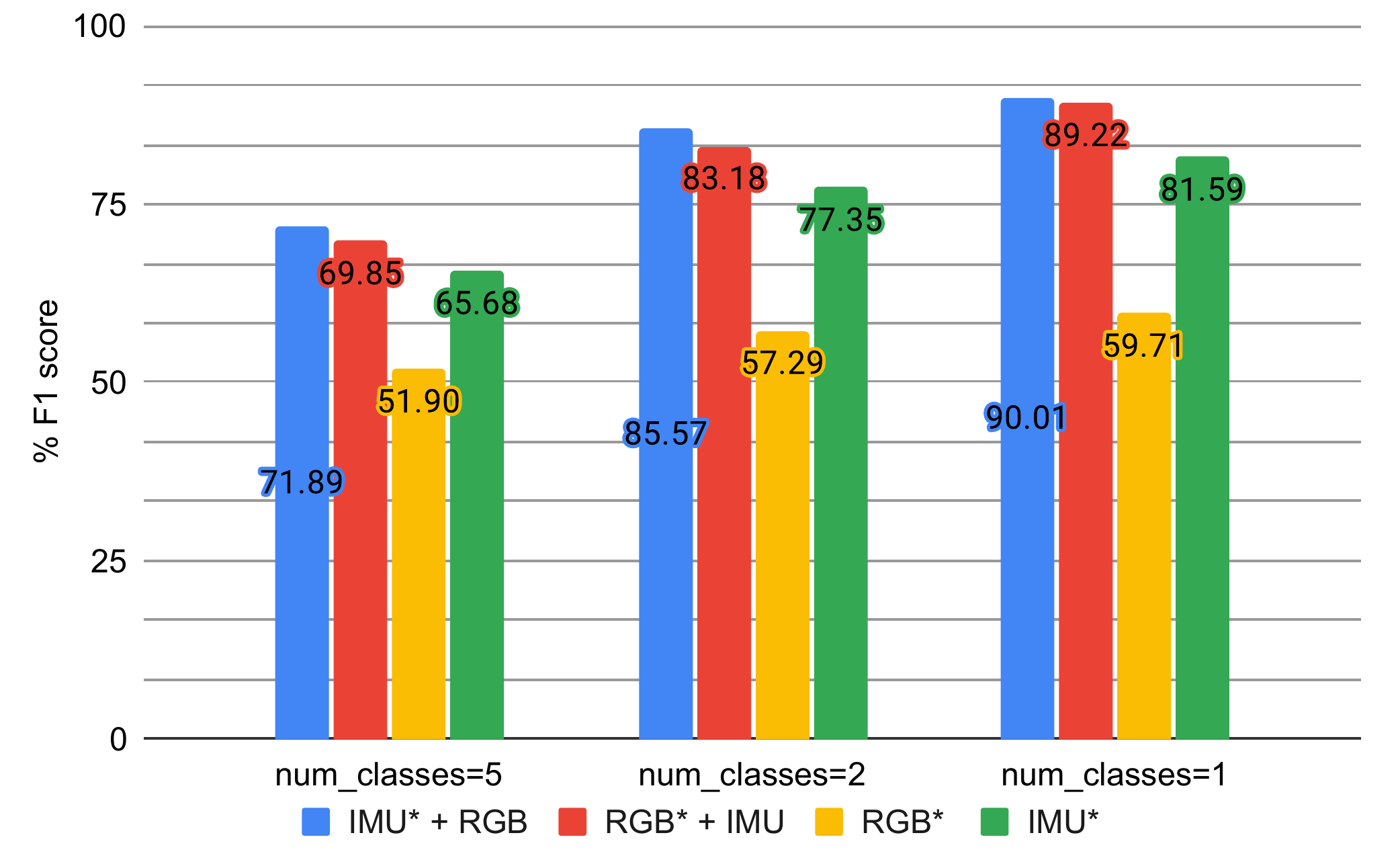}
\end{minipage}
\caption{Zero shot performances on UTD-MHAD dataset. * indicates the input modality from which the action classes were hidden. The four colour bars indicates  different input modality combinations and the three columns represent performances when num\_classes=\textit{$<x>$} were hidden from the input. \textit{Left:} shows the \% accuracy, while \textit{Right:} shows the \% F1 score.} 
\label{fig:zero_shot_performance}
 \vspace{-0.2in}
\end{figure*}

\vspace{-0.1in}
We also test our framework on zero shot setting, and compare against zero shot on RGB and IMU (When few actions classes in the data are hidden during pre-training). During the fine-tuning, the action classes stay hidden to ensure that no data leak happens during the training.
We compare the performances on the test dataset for when single modalities are evaluated for zero shot, and when one of the modalities have seen the action classes in Figure~\ref{fig:zero_shot_performance} in multi-modal setting.
For single modality, we see a performance drop of 2\% when one action class is hidden and a drop of 14\% when 5 action classes are hidden for the IMU data. Whereas, for video, we see a larger drop of 4\% when one class is hidden and 12\% when 5 classes are hidden. 
The multi-modal setup also sees a drop in performance, but performs significantly better to their single modal counterparts. 
We further notice that when action classes are hidden from IMU data input, but are available in the video input, ($IMU^*+RGB$) performs better than ($RGB^*+IMU$) setup. 
This validates our hypothesis that RGB videos are better at providing complementing information to the model when the IMU input lacks it. 

\vspace{-0.15in}
\section{Conclusion}
\vspace{-0.05in}
\label{sec:conclusion}

We have explored multimodal HAR using IMU signals and RGB videos, and the role of pre-training feature encoders for multimodal HAR in this project. We concluded that multimodal HAR outperforms single modal HAR significantly for all datasets.
Furthermore, training in two-stages helps the individual encoders to extract relevant information which when fused performs better than other SOTA works.
We show the effectiveness of the multi-modal training when smaller datasets are used in training, and also in a zero-shot setting.
For the next steps of this project, we would like to explore the effectiveness of using self-attention in learning joint feature representation from multiple modalities and learn better feature fusing protocols, and also explore other modalities for efficient HAR systems.

\vspace{-0.1in}
{
\small
\bibliographystyle{IEEE}
\bibliography{bib}
}
\clearpage
\end{document}